\documentclass{article}

\usepackage{microtype}
\usepackage{graphicx}
\usepackage{subcaption}
\usepackage{booktabs} 
\usepackage{multirow}
\usepackage{multicol}

\usepackage{fancyvrb}
\usepackage{hyperref}

\usepackage{tikz}
\usetikzlibrary{arrows.meta,positioning,fit,backgrounds,calc}

\definecolor{hlblue}{RGB}{173,216,230}
\definecolor{hlorange}{RGB}{255,200,100}

\newcommand{\tok}[2][]{%
  \ifx&#1&\texttt{#2}\else
    \tikz[baseline=(X.base)]{\node[fill=#1,draw=#1!60!black,rounded corners=2pt,inner sep=1.5pt,font=\ttfamily](X){#2};}%
  \fi
}

\usepackage[accepted]{icml2026} 

\usepackage{amsmath}
\usepackage{amssymb}
\usepackage{mathtools}
\usepackage{amsthm}

\icmltitlerunning{In-Context Learning Amplifies a Latent Symbolic Circuit}

\begin{document}

\twocolumn[
  \icmltitle{In-Context Learning Amplifies a Latent Symbolic Circuit}

  \begin{icmlauthorlist}
    \icmlauthor{Melissa Wessel}{ind}
  \end{icmlauthorlist}

  \icmlaffiliation{ind}{Independent Researcher}

  \icmlcorrespondingauthor{Melissa Wessel}{melissawessel00@gmail.com}

  \icmlkeywords{mechanistic interpretability, in-context learning, function vectors, circuit analysis}

  \vskip 0.3in
]



\printAffiliationsAndNotice{}  

\begin{abstract}
  Large language models can learn abstract rules from just a few in-context examples, but how their internal mechanisms activate as examples accumulate is not well understood. We trace a three-stage symbolic reasoning circuit (abstraction, induction, retrieval) across shot counts in three model families and find it is detectable and functional well before the model achieves high accuracy. Per-head causal contribution grows up to $8\times$ from 1- to 10-shot, and cross-shot activation patching raises accuracy from 1\% to 56\% at 0-shot and 17\% to 88\% at 1-shot. Function vectors scaled and injected at 0-shot rescue accuracy up to 86\%, largely substituting for the induction stage but depending critically on an intact downstream retrieval stage. The infrastructure for abstract rule-following is present in the weights before any demonstrations; in-context examples, function vectors, and related interventions appear to supply input to the same latent circuit.
\end{abstract}

\section{Introduction}
Large language models have demonstrated striking capabilities on abstract reasoning tasks, including those requiring generalization over arbitrary tokens \citep{webb2023emergent, musker2025llms}, but the mechanisms underlying these capabilities are not yet well understood. Recent work has identified a three-stage circuit for symbolic reasoning \cite{yang2025emergentsymbolicmechanismssupport}; that work, however, leaves open the question of what role in-context examples play in the circuit's operation. More broadly, the relationship between in-context demonstrations and internal mechanisms remains an active area of study: work on function vectors and concept vectors has characterized how task identity is represented during in-context learning (ICL, \citealp{todd2024function, hendel2023context, opielka2026causality}), while theoretical analyses have modeled ICL as implicit algorithm construction within the forward pass \cite{vonoswald2023transformers, akyurek2024incontext}. A priori, several options are possible: new heads could be recruited into the circuit as examples accumulate, entire stages could come online discretely as with induction heads during pretraining \cite{elhage2021mathematical, olsson2022incontextlearninginductionheads}, or heads important at low shot counts could become irrelevant at high shot counts as the model shifts strategy.  In this work, we trace the developmental trajectory of the \citeauthor{yang2025emergentsymbolicmechanismssupport} circuit across shot counts and find that it is identifiable and functional well before the model achieves high accuracy, suggesting that even in abstract configurations of arbitrary tokens absent any semantic meaning, examples amplify a latent computation.

We study the circuit identified by \citet{yang2025emergentsymbolicmechanismssupport} for algebraic identity rule induction \cite{Marcus1999}, a task in which models must generalize over arbitrary tokens, precluding reliance on memorized semantic associations \citep{webb2023emergent, musker2025llms}. The model sees N in-context examples of three-token patterns (ABC, delimited by \^{} , separated by line breaks) and must complete the final token of a partial example. Two rules are used: ABA (C = A) and ABB (C = B). A sample 2-shot ABA prompt:

\begin{center}
\begin{BVerbatim}
iac ^ ilege ^ iac
ptest ^ yi ^ ptest
ks ^ ixe ^ __?
\end{BVerbatim}
\end{center}

The circuit comprises three stages: Symbolic Abstraction (SA) heads convert input tokens into variables, Symbolic Induction (SI) heads perform sequence induction over those variables, and Retrieval (Ret) heads recover the predicted token. \citet{yang2025emergentsymbolicmechanismssupport} observed the circuit across several architecture families, however, their analysis focused on the mature circuit at high accuracy, primarily showcasing Llama 3.1-70B \cite{grattafiori2024llama3herdmodels} at 2-shot where accuracy already exceeds 95\%. By contrast, smaller models such as Gemma 2-2B \cite{gemmateam2024gemma2improvingopen} require up to 10 shots to reach comparable performance, providing a natural window into how the circuit develops as in-context examples accumulate.

We trace this development in Gemma 2-2B \cite{gemmateam2024gemma2improvingopen}, Llama 3.1-8B \cite{grattafiori2024llama3herdmodels} and Qwen 3-4B \cite{yang2025qwen3technicalreport} and find a consistent picture: stable circuit topology with quantitative signal growth. The circuit is observable via causal mediation analysis (CMA) from just 1-shot, when baseline accuracy for Gemma 2-2B is only 17\%, with per-head causal contribution increasing up to $8\times$ between 1-shot and 10-shot. Patching 10-shot activations of SA, SI, and Retrieval heads into 1-shot prompts raises accuracy to 88\%. At 0-shot, SA heads cannot be tested as they write to the C positions which do not exist, however the downstream SI and Retrieval stages are functional and writable, with 10-shot activations raising accuracy from 1\% to 56\%, activating infrastructure that predates any demonstrations. SI stage patches specifically transfer even across prompts with no shared tokens, paralleling prior findings on function vectors \citep{todd2024function}. Function vectors extracted from CMA-significant heads rescue 0-shot accuracy to 86\% when injected into the residual stream but collapse to 13\% when the downstream Retrieval stage heads are ablated: the FV is causally interchangeable with the SI stage's output (fully on ABB, partially on ABA), contingent on Ret remaining available to read it out.

These results contribute to a mechanistic account of ICL of abstract rules: examples strengthen signal through circuitry that already exists in the weights. ICL has been studied extensively at behavioral \citep{brown2020language} and theoretical \citep{vonoswald2023transformers, akyurek2024incontext} levels; our results address the mechanistic question of how examples reshape computation in the forward pass, at least for this task family. They also speak to the relationship between ICL and interventions such as function vectors \citep{todd2024function} and activation steering \citep{zou2023representation, turner2024steeringlanguagemodelsactivation, liu2024incontextvectorsmakingcontext}, suggesting that such interventions may succeed in part because they supply input signal into pre-existing computational pathways. 

\begin{figure}[t]
    \centering
    \includegraphics[width=0.5\textwidth]{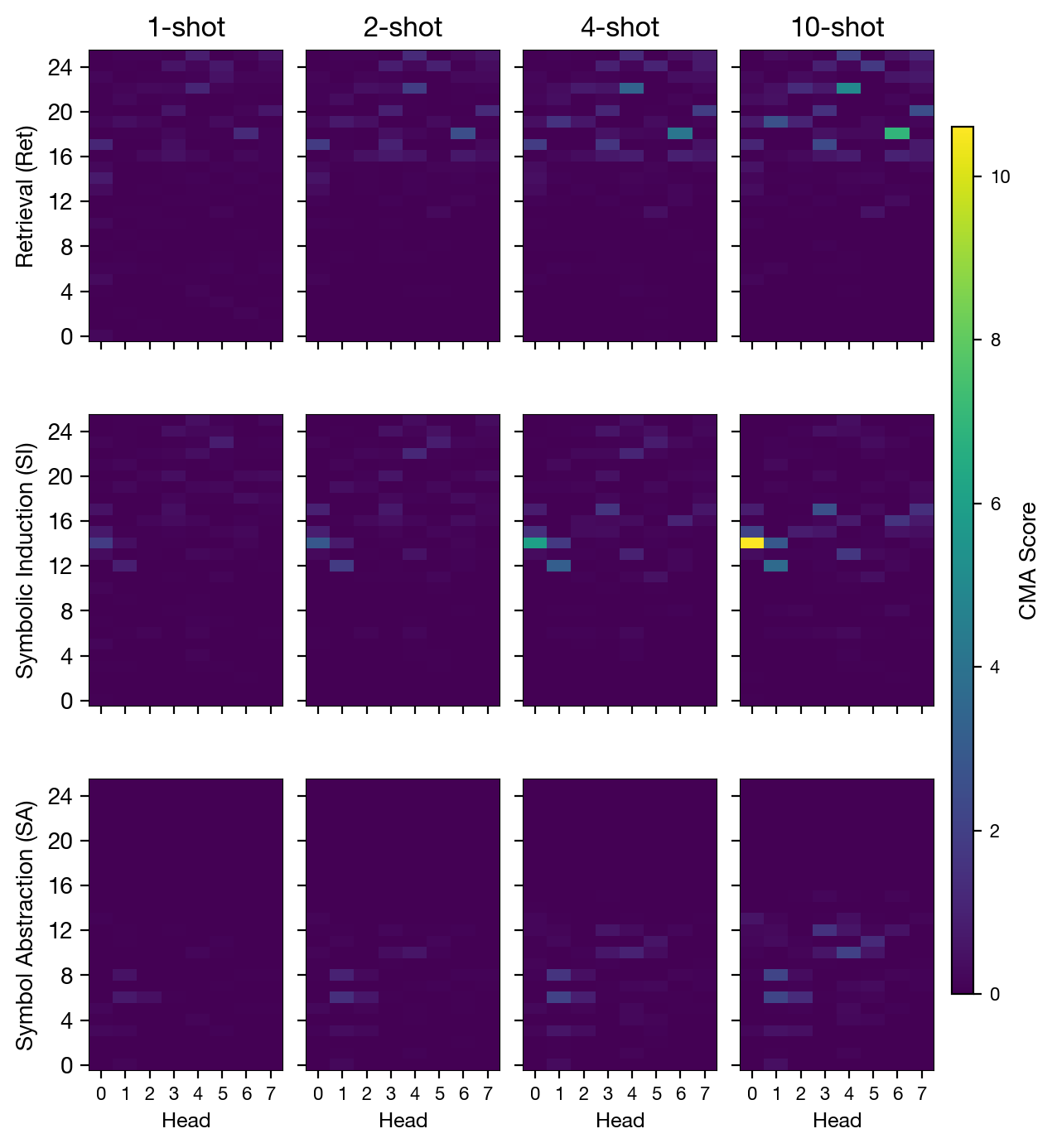}
    \caption{\textbf{The Symbolic Reasoning Circuit in Gemma 2-2B at 1 through 10-Shot.} CMA Scores by layer and head by circuit stage across shot counts.}
    \label{fig:circuit-by-shot-gemma-2-2b}
\end{figure}

\begin{figure*}[t]
  \centering
  \begin{subfigure}[b]{0.48\textwidth}
    \centering
    \includegraphics[width=\textwidth]{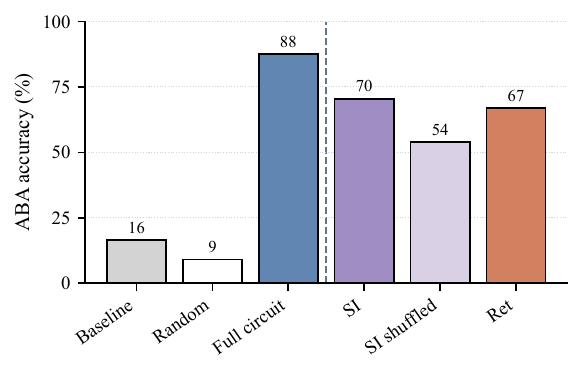}
    \caption{1-Shot}
    \label{fig:one-shot-patch}
  \end{subfigure}
  \hfill
  \begin{subfigure}[b]{0.48\textwidth}
    \centering
    \includegraphics[width=\textwidth]{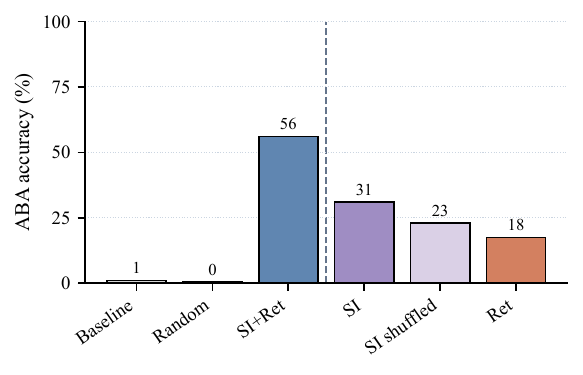}
    \caption{0-Shot}
    \label{fig:zero-shot-patch}
  \end{subfigure}
  \caption{\textbf{Cross-Shot Patching Results (ABA).} Accuracy of Gemma 2-2B when patching 10-shot donor activations into 1-shot (a) and 0-shot (b) target prompts. Full results for all models and both rules are in Appendix Figures~\ref{fig:appendix_patching_gemma_2_2b}--\ref{fig:appendix_patching_qwen3_4b}.}
  \label{fig:cross-shot-patching-results}
\end{figure*}

\section{Circuit Amplification}
In Gemma 2-2B, 1-shot accuracy is 17\% and increases to 93\% by 4-shot, reaching 99\% by 10-shot. Llama 3.1-8B performs better but is still imperfect at 1-shot (60\% ABA), and Qwen 3-4B tracks Gemma more closely (27\% ABA, 55\% ABB at 1-shot). We evaluate whether the three-stage topology \citet{yang2025emergentsymbolicmechanismssupport} identified is observable from 1-shot across all three models, and track how it changes from 1 to 10 shots.

\paragraph{Method.}
We follow \citet{yang2025emergentsymbolicmechanismssupport} and apply their CMA from 0 to 10-shot. Their method uses two contrastive conditions. The abstract condition (varying rule, fixed answer token) identifies SA heads when patched at C positions and SI heads when patched at the query position. The token condition (fixed rule, varying answer token) identifies Retrieval heads at the query position. To quantify topology stability across shot counts we rank heads per stage by mean CMA score and compare rankings using rank-biased overlap (RBO; \citealt{webber2010rbo}, $p{=}0.9$), a top-weighted [0,1] similarity. 

\paragraph{Results.}
The three-stage circuit is observable from 1-shot in all three models, and changes across shot counts are dominated by signal growth rather than topology change. We describe Gemma 2-2B in detail as it shows the largest accuracy transition, then note how Llama 3.1-8B and Qwen 3-4B compare.

In Gemma, CMA scores grow up to $8\times$ from 1 to 10 shots without major changes in circuit topology (Figure~\ref{fig:circuit-by-shot-gemma-2-2b}). The top-5 heads per stage overlap 4-5/5 across adjacent shot counts for SI and Retrieval (RBO 0.81-0.94; random baseline 0.12, Table~\ref{tab:rbo}), and 3-5/5 for SA (RBO 0.66-0.91), consistent with SA's lower overall CMA signal. The persistent SA core (L6H1, L6H2, L8H1, Table~\ref{tab:persistent-core}) is present from 1-shot onward, and no stage exhibits discrete onset. Signal grows within this relatively fixed topology: the dominant SI head L14H0 increases from 0.9 to 7.4 ($8.1\times$) for ABB, and the top Retrieval head L18H6 from 1.3 to 7.0 ($5.6\times$), with amplification largest across the 2-to-4-shot transition where accuracy itself jumps most sharply. The same pattern holds on letter-string analogies (e.g., [a b c] $\rightarrow$ [a b d], [d e f] $\rightarrow$ [d e ?, see Appendix~\ref{appendix:lsa}).

Llama 3.1-8B shows a strikingly parallel pattern despite its different accuracy regime (60\% $\rightarrow$ 100\%): 4/5 persistent core SI heads, 4/5 Retrieval, and 3/5 SA. Qwen 3-4B is noisier at top-5 (1/5 SA, 1/5 SI, 3/5 Ret) but recovers at top-10 (5/10 SA, 4/10 SI, 7/10 Ret); inspection of the rankings shows the top-5 noise reflects reranking among the leading heads rather than new heads entering. Stability here is a claim about which heads carry the computation, not exact rank order; the lowest-signal regimes (Qwen at 1-shot, SA broadly) show the most reranking among the leading heads.

\section{Cross-Shot Activation Patching}
CMA reveals that the circuit is observable at low shot counts, but observation alone cannot establish that it is functional and similar enough to its more mature counterparts for its outputs to be transferrable. To test this, we intervene directly: patching activations from 10-shot forward passes into 1-shot and even 0-shot prompts and measuring the effect on accuracy.

\subsection{Patching from 10-shot to 1-shot}
For each 1-shot target, we construct a matched pair where the target's single example is the final example of the 10-shot donor (shared tokens, same rule). We patch at stage-specific positions: SA heads at C positions, SI and Ret heads at the query position. We test each stage independently and all stages together, including a random control and a shuffled token control: the donor uses the same abstract rule but entirely different tokens. We evaluate 200 prompts per rule.

Patching all stages from the 10-shot donor rescues ABA accuracy from 17\% to 88\% (Figure~\ref{fig:one-shot-patch}), and ABB from 57\% to 93\%. Patching Retrieval heads alone yields only 67\% (ABA), confirming that the rescue depends on upstream rule computation. Pure answer-token injection would predict the model's native 10-shot ceiling (99\%); the gap below that implies upstream rule computation.

\subsection{Patching from 10-shot to 0-shot}
For 0-shot targets, the prompt contains only the query \texttt{A\^{}B\^{}} with no in-context examples. We construct matched triplets: a 10-shot ABA donor, a 10-shot ABB donor, and a 0-shot prompt sharing the same query tokens (Figure~\ref{fig:cross-shot-patching}). Using permutation-significant heads from 10-shot CMA (SI and Ret only, since SA writes to C positions absent at 0-shot), we patch per-head attention output from each donor's query position to the 0-shot forward pass.

At 0-shot, rule-following accuracy is 1\% (ABA) and 8\% (ABB), yet patching 7 SI and 9 Retrieval heads out of 208 total rescues accuracy to 56\% (ABA, Figure~\ref{fig:zero-shot-patch}) and 73\% (ABB). Patching Ret heads alone yields an accuracy of 18\% (ABA), again ruling out simple answer injection by the retrieval stage. Critically, the same \texttt{A\^{}B\^{}} prompt predicts A when patched from an ABA donor and B from an ABB donor; the injected activations, not the prompt, determine the answer. Patching random non-circuit heads yields $<$1\% (ABA), confirming the effect is specific to the three-stage circuit.

\subsection{Token Invariance of the SI Stage}

SI heads appear to carry abstract, largely token-invariant rule information, though the degree of token-invariance varies by model. At 1-shot, matched-token and shuffled-token SI patching rescue Llama 3.1-8B accuracy to nearly identical levels on both rules (ABA: 83\% vs 83\%; ABB: 83\% vs 81\%), despite donor and target sharing no tokens. Gemma 2-2B and Qwen 3-4B show only partial transfer: shuffled-token SI patching captures roughly two-thirds (69\%) of the matched-token effect on ABA in Gemma and roughly half (52\%) in Qwen.

The same pattern holds at 0-shot. In Llama 3.1-8B, shuffled-token SI patching is indistinguishable from matched-token on both rules (ABA: 36\% vs 39\%; ABB: 37\% vs 38\%). In Gemma 2-2B, shuffled patching captures 73\% of the matched-token ABA rescue. For Qwen 3-4B we report ABB rather than ABA: the 0-shot ABA condition produces only a 7-point SI-only effect over a 0\% baseline, making the matched-vs-shuffled ratio uninformative; on ABB, shuffled patching captures 49\% of the matched-token rescue. SI heads thus encode rule information in a largely token-abstract form that is stable across shot counts; the residual matched-prompt advantage in Gemma and Qwen suggests an additional token-specific component layered on the abstract one.

\section{Function Vectors as Circuit Signal}

If examples amplify signal through a latent circuit, function vectors \citep{todd2024function} should be readable as one instance of that signal. \citet{yang2025emergentsymbolicmechanismssupport} report that FV-producing heads and SI heads largely overlap ($r{=}0.82$) and conjecture FVs implement symbolic induction. We test this causally.

Following \citet{todd2024function}, we extract $z\,W_O$ at the last query position from correct 10-shot Gemma 2-2B passes, average across prompts, and sum across the 29 permutation-test-significant heads at 10-shot to obtain one vector per rule. We add $s \cdot \mathrm{FV}_r$ to the residual stream at layer 12, last query position, of held-out 0-shot prompts, then re-run with the 7 SI heads or 9 Ret heads zero-ablated at the query position. We use $s{=}25$ and layer 12, where rescue plateaus for both rules (full sweep in Appendix~\ref{appendix:fv}).

\begin{figure}[t]
    \centering
    \includegraphics[width=\columnwidth]{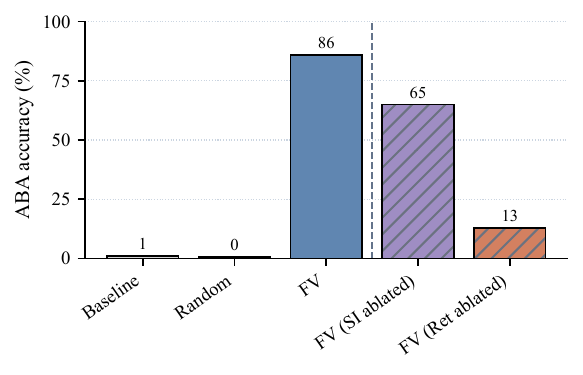}
    \caption{\textbf{Function-vector rescue at 0-shot, ABA.}}
    \label{fig:fv}
  \end{figure}

FV injection rescues 0-shot accuracy from 1\% to 86\% on ABA (Figure~\ref{fig:fv}) and from 11\% to 89\% on ABB, with norm-matched random vectors collapsing accuracy to 0.5\%, ruling out generic perturbation as the source of rescue. Ret ablation at the query position drops the rescue to 13\% (ABA) and 29\% (ABB); the rule signal is supplied but cannot be read out. SI ablation is rule-asymmetric: it reduces the ABA rescue to 65\% but leaves ABB unchanged at 89\%. The FV appears to fully substitute for SI's output on ABB and partially on ABA, while Ret's role as readout cannot be bypassed by injection upstream of it.

\section{Discussion}
Our results suggest that in-context learning of abstract rules operates predominantly through signal amplification within a latent circuit. The same three-stage topology is identifiable from 1-shot, when accuracy is just 17\%, its core heads persist across shot counts, and cross-shot patching confirms the circuit is functional when supplied with sufficient input signal. The FV result reinforces this from another angle: a single static vector roughly substitutes for the SI stage at 0-shot, but only when Ret is available to read it out. On this view, demonstrations are better understood as input to a latent computation than as specifications for one constructed in the forward pass.

The partial token-invariance of SI patches and the success of FV injection both connect to prior work on task, function, and concept vectors \cite{hendel2023context, todd2024function, opielka2026causality}, which show that ICL task identity can be captured in transferable activation patterns. Our FV result sharpens the relationship: the vector is not merely correlated with SI heads, as observed by \citeauthor{yang2025emergentsymbolicmechanismssupport}, but causally nearly interchangeable with that stage's output, contingent on Ret remaining available to consume it. This frames such transferable representations as readouts of pre-existing computational pathways more so than freestanding mechanisms. The same logic extends naturally to activation steering \cite{zou2023representation}, where fixed perturbations reliably shift behavior.

Several limitations apply. Contrastive CMA isolates only rule-discriminating components, potentially missing early-layer infrastructure heads that support both rules and likely explains the residual gap between patched (88\%) and native 10-shot (99\%) accuracy. The task, while requiring generalization over arbitrary tokens, has a simple relational structure (essentially a 3-column table with a copy operation); the amplification pattern also holds on letter-string analogies (which involve no copying), but whether it generalizes to richer abstract reasoning is open.

A broader implication: if in-context learning of this rule operates by amplifying signal through a pre-existing circuit, then understanding ICL behavior reduces in part to understanding what circuits exist in the weights and what supplies their input. Open questions about what makes a task learnable in-context, why some interventions transfer while others don't, and why context is sometimes used poorly even when present then become questions about latent circuit inventory and accessibility.

\section*{Impact Statement}

This paper presents work whose goal is to advance the field of Machine Learning. There are many potential societal consequences of our work, none which we feel must be specifically highlighted here.

\bibliography{paper}
\bibliographystyle{icml2026}

\newpage
\appendix
\onecolumn
\section{Appendix}

\subsection{Code and Hardware}

All code is written in Python using TransformerLens and the HuggingFace Transformers library. Gemma 2-2B experiments were run on a 36GB Apple Silicon MacBook Pro (MPS); Llama 3.1-8B and Qwen 3-4B experiments were run on a single NVIDIA A40 GPU. All model weights are loaded in the \texttt{bfloat16} format. Code and bundled paper results are available at \url{https://github.com/melissawessel/icl-symbolic}.

\subsection{Experimental Details}

\subsubsection{Models}

\begin{table}[h]
\centering
\small
\begin{tabular}{lcccccc}
\toprule
Model & Layers & Q heads & KV heads & $d_{\text{model}}$ & $d_{\text{head}}$ & BOS \\
\midrule
Gemma 2-2B \cite{gemmateam2024gemma2improvingopen}        & 26 & 8  & 4 & 2304 & 256 & yes \\
Llama 3.1-8B \cite{grattafiori2024llama3herdmodels}        & 32 & 32 & 8 & 4096 & 128 & yes \\
Qwen 3-4B \cite{yang2025qwen3technicalreport}              & 36 & 32 & 8 & 2560 & 128 & no  \\
\bottomrule
\end{tabular}
\caption{Model architectures. All three use grouped-query attention (GQA). All experiments use \texttt{bfloat16} precision and load the models via TransformerLens, which materializes one $W_K$/$W_V$ per query head so attention-head indexing is uniform across architectures.}
\label{tab:models}
\end{table}
\subsubsection{Prompts}

Following \citet{yang2025emergentsymbolicmechanismssupport}, tokens are sampled from a 72K-entry English-token vocabulary released with their codebase, restricted at runtime to entries that (a) encode to a single token under the model's tokenizer and (b) do not merge with the separator \texttt{\^{}}. Each in-context example occupies six tokens (\texttt{A \^{} B \^{} C \textbackslash n}) and the query occupies four (\texttt{A \^{} B \^{}}). With BOS offset $b \in \{0,1\}$, example $i \in \{0,\ldots,n{-}1\}$ has its $C$ position at $b + 6i + 4$ and the final query position lies at $b + 6n + 3$. Total prompt length is $b + 6n + 4$ tokens. Each prompt is validated to tokenize to exactly this length before use.

Accuracy is evaluated on $N{=}500$ prompts per rule per shot count (1000 total per shot count). CMA uses $N{=}200$ matched context pairs per condition (100 per rule direction). Cross-shot patching uses $N{=}200$ matched donor/target pairs per rule. Random seeds are unique per (model, $n$, rule) so token identity does not co-vary with shot count.

Matched pairs are constructed as follows. For the \emph{abstract} CMA condition, a base ABA prompt $a_1\,b_1\,a_1 \mid \ldots \mid a_q\,b_q\,?$ is paired with an exp prompt that swaps the first two tokens of every example and the query ($b_1\,a_1\,a_1 \mid \ldots \mid b_q\,a_q\,?$). Both prompts share their final tokens but differ in which abstract role each token plays; the same construction inverted for ABB. For the \emph{token} condition the rule is held fixed and the query tokens are swapped, isolating the literal-token component. For cross-shot patching, the target's $n_t$ examples are the \emph{last} $n_t$ examples of the matched $n_d$-shot donor (same tokens, same rule). The 0-shot triplet construction pairs an empty-context query \texttt{A\^{}B\^{}} with two 10-shot donors (one ABA, one ABB) sharing the query tokens but with otherwise disjoint demonstrations.

\subsubsection{Causal Mediation Analysis}

CMA follows \citet{yang2025emergentsymbolicmechanismssupport}, Algorithm~1. For each context pair (base $c_2$, exp $c_1$) and each attention head $(\ell, h)$, we run the base context forward, cache the head's per-token output \texttt{hook\_z} at the patch position, and run the exp context with that head's activation overwritten at the patch position. The score is the change in logit difference,
\begin{equation*}
\text{CMA}(\ell, h) \;=\; \big[\,\mathcal{L}(c_1^*; a_{\text{causal}}, a_{\text{exp}})\,\big] - \big[\,\mathcal{L}(c_1; a_{\text{causal}}, a_{\text{exp}})\,\big],
\end{equation*}
where $\mathcal{L}(c; a_1, a_2) = \text{logit}(a_1 \mid c) - \text{logit}(a_2 \mid c)$, $c_1^*$ denotes the patched exp context, $a_{\text{causal}}$ is the answer the patched model should predict if the head carries the rule, and $a_{\text{exp}}$ is the answer of the unpatched exp context. Per-pair scores are stored as $[N_{\text{valid}}, n_{\text{layers}}, n_{\text{heads}}]$ and averaged across pairs and across the two rule directions for ranking and significance.

\begin{table}[!ht]
\centering
\small
\begin{tabular}{lll}
\toprule
Stage & Patch position(s) & CMA condition \\
\midrule
Symbolic Abstraction (SA) & all $C$ positions: $\{b + 6i + 4 : 0 \le i < n\}$ & abstract \\
Symbolic Induction (SI)   & final query position: $b + 6n + 3$            & abstract \\
Retrieval (Ret)           & final query position: $b + 6n + 3$            & token    \\
\bottomrule
\end{tabular}
\caption{Patch positions per stage. SA writes to the $C$ tokens of the demonstrations (so no SA evaluation is possible at $n{=}0$). SI and Retrieval write to the same query position but are distinguished by the contrastive condition.}
\label{tab:cma-positions}
\end{table}

We use two pair-validity filters: \texttt{threshold} (keep pairs where both contexts assign $\Pr(\text{answer}) \ge 0.9$, used by \citeauthor{yang2025emergentsymbolicmechanismssupport} on high-accuracy models) and \texttt{correct} (keep pairs where the argmax matches the answer). Below 4-shot, Gemma and Qwen rarely clear the 0.9 probability threshold even on prompts they answer correctly, so we use the \texttt{correct} filter at all shot counts to keep the methodology comparable across models and shot counts. The number of valid pairs after filtering is reported alongside each result.

\begin{figure*}[t]
    \centering
    \resizebox{0.95\textwidth}{!}{



\begin{tikzpicture}[
  box/.style={draw,rounded corners=4pt,fill=gray!8,inner sep=6pt,align=center,font=\small},
  lbl/.style={font=\sffamily\bfseries\small,anchor=west},
  arr/.style={-{Stealth[length=4pt]},thick},
  patch/.style={-{Stealth[length=4pt]},gray,thick},
  outp/.style={font=\ttfamily\small},
  colhdr/.style={font=\sffamily\bfseries\small,anchor=south},
  node distance=8mm and 12mm,
]

\def\colA{1.5}    
\def\colB{7.3}    
\def\colC{14}     
\def\colD{19.8}   

\def\rowBase{0}
\def\rowDonor{-3.2}
\def\rowTarget{-6.5}

\node[colhdr] at ($({\colA},2.5)!0.5!({\colB},0.6)$) {1-shot};
\node[colhdr] at ($({\colC},2.5)!0.5!({\colD},0.6)$) {0-shot};
\draw[gray,thin] ({(\colB+\colC)/2+0.5},1.0) -- ({(\colB+\colC)/2+0.5},{\rowTarget-1.2});

\node[lbl] at (-2.5,{\rowBase}) {Baseline};
\node[lbl] at (-2.5,{\rowDonor}) {Donor};
\node[lbl] at (-2.5,{\rowTarget}) {Target};

\node[box] (b1a) at (\colA,\rowBase) {
  \tok{ptest}\,\tok{\^{}}\,\tok{yi}\,\tok{\^{}}\,\tok{ptest}\\[2pt]
  \tok{ks}\,\tok{\^{}}\,\tok{ixe}\,\tok{\^{}}
};
\node[outp,right=8mm of b1a] (b1a-out) {??};
\draw[arr] (b1a) -- (b1a-out);

\node[box] (d1a) at (\colA,\rowDonor) {
  \tok{iac}\,\tok{\^{}}\,\tok{ilege}\,\tok{\^{}}\,\tok{iac}\\[2pt]
  $\vdots$\\[2pt]
  \tok{ptest}\,\tok{\^{}}\,\tok{yi}\,\tok{\^{}}\,\tok[hlorange]{ptest}\\[2pt]
  \tok{ks}\,\tok{\^{}}\,\tok{ixe}\,\tok[hlblue]{\^{}}
};
\node[outp,right=8mm of d1a] (d1a-out) {ks};
\draw[arr] (d1a) -- (d1a-out);

\node[box] (t1a) at (\colA,\rowTarget) {
  \tok{ptest}\,\tok{\^{}}\,\tok{yi}\,\tok{\^{}}\,\tok[hlorange]{ptest}\\[2pt]
  \tok{ks}\,\tok{\^{}}\,\tok{ixe}\,\tok[hlblue]{\^{}}
};
\node[outp,right=8mm of t1a] (t1a-out) {ks};
\draw[arr] (t1a) -- (t1a-out);

\draw[patch] (d1a) -- node[right,font=\scriptsize\sffamily,text=gray]{patch} (t1a);

\node[box] (b1b) at (\colB,\rowBase) {
  \tok{ptest}\,\tok{\^{}}\,\tok{yi}\,\tok{\^{}}\,\tok{yi}\\[2pt]
  \tok{ks}\,\tok{\^{}}\,\tok{ixe}\,\tok{\^{}}
};
\node[outp,right=8mm of b1b] (b1b-out) {??};
\draw[arr] (b1b) -- (b1b-out);

\node[box] (d1b) at (\colB,\rowDonor) {
  \tok{iac}\,\tok{\^{}}\,\tok{ilege}\,\tok{\^{}}\,\tok{ilege}\\[2pt]
  $\vdots$\\[2pt]
  \tok{ptest}\,\tok{\^{}}\,\tok{yi}\,\tok{\^{}}\,\tok[hlorange]{yi}\\[2pt]
  \tok{ks}\,\tok{\^{}}\,\tok{ixe}\,\tok[hlblue]{\^{}}
};
\node[outp,right=8mm of d1b] (d1b-out) {ixe};
\draw[arr] (d1b) -- (d1b-out);

\node[box] (t1b) at (\colB,\rowTarget) {
  \tok{ptest}\,\tok{\^{}}\,\tok{yi}\,\tok{\^{}}\,\tok[hlorange]{yi}\\[2pt]
  \tok{ks}\,\tok{\^{}}\,\tok{ixe}\,\tok[hlblue]{\^{}}
};
\node[outp,right=8mm of t1b] (t1b-out) {ixe};
\draw[arr] (t1b) -- (t1b-out);

\draw[patch] (d1b) -- node[right,font=\scriptsize\sffamily,text=gray]{patch} (t1b);

\node[box] (b0a) at (\colC,\rowBase) {
  \tok{ks}\,\tok{\^{}}\,\tok{ixe}\,\tok{\^{}}
};
\node[outp,right=8mm of b0a] (b0a-out) {??};
\draw[arr] (b0a) -- (b0a-out);

\node[box] (d0a) at (\colC,\rowDonor) {
  \tok{iac}\,\tok{\^{}}\,\tok{ilege}\,\tok{\^{}}\,\tok{iac}\\[2pt]
  $\vdots$\\[2pt]
  \tok{ptest}\,\tok{\^{}}\,\tok{yi}\,\tok{\^{}}\,\tok{ptest}\\[2pt]
  \tok{ks}\,\tok{\^{}}\,\tok{ixe}\,\tok[hlblue]{\^{}}
};
\node[outp,right=8mm of d0a] (d0a-out) {ks};
\draw[arr] (d0a) -- (d0a-out);

\node[box] (t0a) at (\colC,\rowTarget) {
  \tok{ks}\,\tok{\^{}}\,\tok{ixe}\,\tok[hlblue]{\^{}}
};
\node[outp,right=8mm of t0a] (t0a-out) {ks};
\draw[arr] (t0a) -- (t0a-out);

\draw[patch] (d0a) -- node[right,font=\scriptsize\sffamily,text=gray]{patch} (t0a);

\node[box] (b0b) at (\colD,\rowBase) {
  \tok{ks}\,\tok{\^{}}\,\tok{ixe}\,\tok{\^{}}
};
\node[outp,right=8mm of b0b] (b0b-out) {??};
\draw[arr] (b0b) -- (b0b-out);

\node[box] (d0b) at (\colD,\rowDonor) {
  \tok{iac}\,\tok{\^{}}\,\tok{ilege}\,\tok{\^{}}\,\tok{ilege}\\[2pt]
  $\vdots$\\[2pt]
  \tok{ptest}\,\tok{\^{}}\,\tok{yi}\,\tok{\^{}}\,\tok{yi}\\[2pt]
  \tok{ks}\,\tok{\^{}}\,\tok{ixe}\,\tok[hlblue]{\^{}}
};
\node[outp,right=8mm of d0b] (d0b-out) {ixe};
\draw[arr] (d0b) -- (d0b-out);

\node[box] (t0b) at (\colD,\rowTarget) {
  \tok{ks}\,\tok{\^{}}\,\tok{ixe}\,\tok[hlblue]{\^{}}
};
\node[outp,right=8mm of t0b] (t0b-out) {ixe};
\draw[arr] (t0b) -- (t0b-out);

\draw[patch] (d0b) -- node[right,font=\scriptsize\sffamily,text=gray]{patch} (t0b);

\def\legendY{\rowTarget-2.0}
\node[anchor=west] at ({(\colA+\colD)/2 - 6}, \legendY) {
  \tikz[baseline=(X.base)]{\node[fill=hlorange,draw=hlorange!60!black,rounded corners=2pt,minimum width=8pt,minimum height=8pt,font=\small](X){\phantom{X}};}
  {\small\sffamily\, SA head patch position (C tokens)}
};
\node[anchor=west] at ({(\colA+\colD)/2 + 1}, \legendY) {
  \tikz[baseline=(X.base)]{\node[fill=hlblue,draw=hlblue!60!black,rounded corners=2pt,minimum width=8pt,minimum height=8pt,font=\small](X){\phantom{X}};}
  {\small\sffamily\, SI \& Retrieval head patch position (query token)}
};

\end{tikzpicture}
}
    \caption{\textbf{Cross-shot activation patching.} Activations from 10-shot donors (middle) are patched into 1-shot (left) and 0-shot (right) targets (bottom), steering outputs to the correct rule-following completion. Yellow highlights indicate the SA head patch position (C tokens); blue indicates the SI and Retrieval head patch position (query token).}
    \label{fig:cross-shot-patching}
\end{figure*}
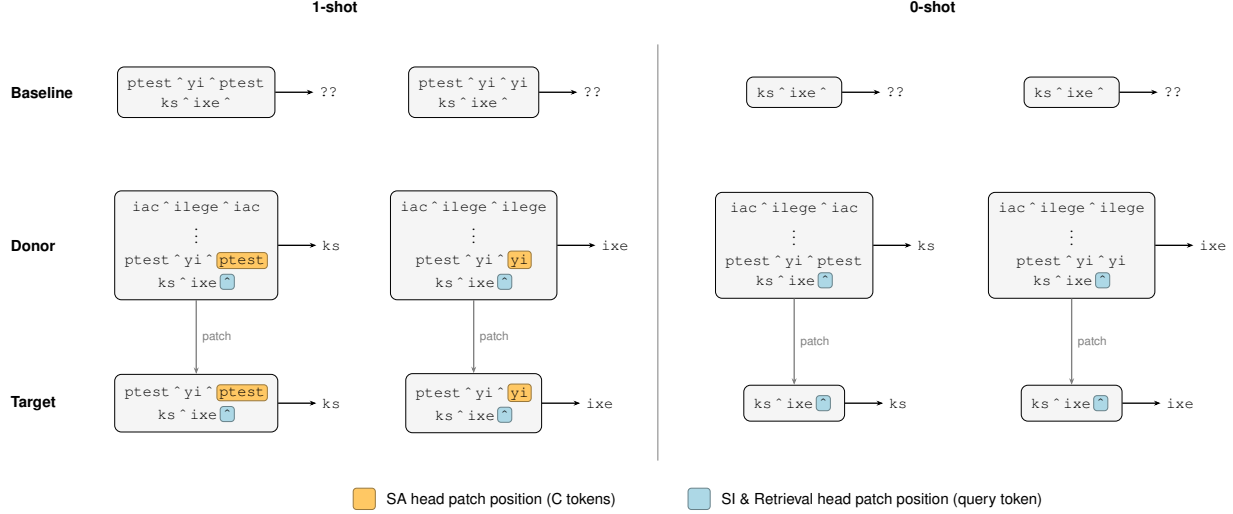

\subsubsection{Activation patching}

All patches operate on \texttt{hook\_z}, the per-query-head attention output before the per-head $W_O$ projection. Patching at this point is per-Q-head rather than per-KV-head, which avoids the GQA complication that arises when patching at $W_V$/$W_K$; TransformerLens materializes one $W_K, W_V$ copy per query head so each query head can be patched independently. Patches replace (not add) the activation: for the patched head at the patched position, the donor's cached \texttt{hook\_z} slice is written into the target's forward pass. All other heads, all other positions, and all MLPs run unmodified. For multi-head patches each head is replaced at its stage-specific position(s).

Two negative controls are used throughout. \emph{Shuffled-token donor:} a donor prompt with the same rule and shot count as the matched donor but a disjoint random token sample. The same heads are patched at the same positions; any rescue that survives must be carried by token-invariant content. \emph{Random non-circuit heads:} a count-matched random sample of attention heads drawn from outside the union of CMA-significant heads, patched at the same positions. The matched-donor rescue must beat both controls for the effect to be attributed to the circuit.

\subsubsection{Significance}

Per-head significance uses a max-statistic permutation test that controls the family-wise error rate across all $(\ell, h)$ positions following \citet{yang2025emergentsymbolicmechanismssupport}, Appendix B.2. For each of $5000$ iterations we sign-flip each pair's score independently with probability $0.5$ (equivalent to permuting the base/exp assignment under the null), recompute the per-head mean, and record the maximum across $(\ell, h)$. The significance threshold is the $95$th percentile of this null distribution; heads whose observed mean exceeds the threshold are reported as significant at $\alpha{=}0.05$.

\subsubsection{Function Vectors}
\label{appendix:fv}

The FV is constructed Todd-style: for the union $\mathcal{H}$ of permutation-significant heads at 10-shot (across both rules; $|\mathcal{H}|{=}29$ for Gemma 2-2B), $\mathrm{FV}_r = \sum_{(\ell,h) \in \mathcal{H}} \mathbb{E}_{c \in \mathcal{C}_r}\big[z^{(\ell,h)}_q(c)\,W_O^{(\ell,h)}\big]$, where $\mathcal{C}_r$ is the set of model-correct 10-shot prompts for rule $r$ and $q$ is the last query position. Injection adds $s \cdot \mathrm{FV}_r$ to \texttt{resid\_pre} at layer $L$, last query position, of held-out 0-shot prompts; ablation zeroes \texttt{hook\_z} for SI or Ret heads at the query position.

We swept $s \in \{1,5,10,25,50,100\}$ and $L \in \{6,8,12,14,16\}$ (200 0-shot prompts per rule, seed 42). Each experiment draws its own 200-prompt evaluation set, so the 0-shot ABB baseline reported here (11.0\%) differs from the cross-shot patching baseline (7.5\%, §3.2) by $\sim$3 pp; this is within the binomial CI half-width ($\approx 4$ pp at $n{=}200$, $p{\approx}0.1$). Rescue saturates around $s{=}25$ across all five layers and collapses at $s{=}100$ (e.g.\ ABA at $L{=}12$: 2.0\%, 22.0\%, 55.0\%, 86.0\%, 84.5\%, 21.5\% for $s = 1, 5, 10, 25, 50, 100$). Per-rule peaks are at $(L{=}8, s{=}50)$ for ABA (88.0\%) and $(L{=}14, s{=}25)$ for ABB (92.0\%); we report $(L{=}12, s{=}25)$, near-peak for both rules (ABA 86.0\%, ABB 88.5\%), so the ablation factorial uses one $(L, s)$ across rules. The full ablation factorial is in Table~\ref{tab:fv-ablation}.

\begin{table}[h]
\centering
\small
\begin{tabular}{lcc}
\toprule
Condition & ABA & ABB \\
\midrule
Baseline (no inject) & 1.0 & 11.0 \\
FV inject & 86.0 & 88.5 \\
FV inject + ablate SI & 65.0 & 89.0 \\
FV inject + ablate Ret & 13.0 & 28.5 \\
Random vector (norm-matched) & 0.0 & 0.5 \\
\bottomrule
\end{tabular}
\caption{FV $\times$ per-stage ablation factorial at $L{=}12$, $s{=}25$. Accuracy (\%), $n{=}200$ per cell.}
\label{tab:fv-ablation}
\end{table}

\subsection{Per-Model CMA Results}

\subsubsection{Llama 3.1-8B}
\begin{figure*}[t]
    \centering
    \includegraphics[width=\textwidth]{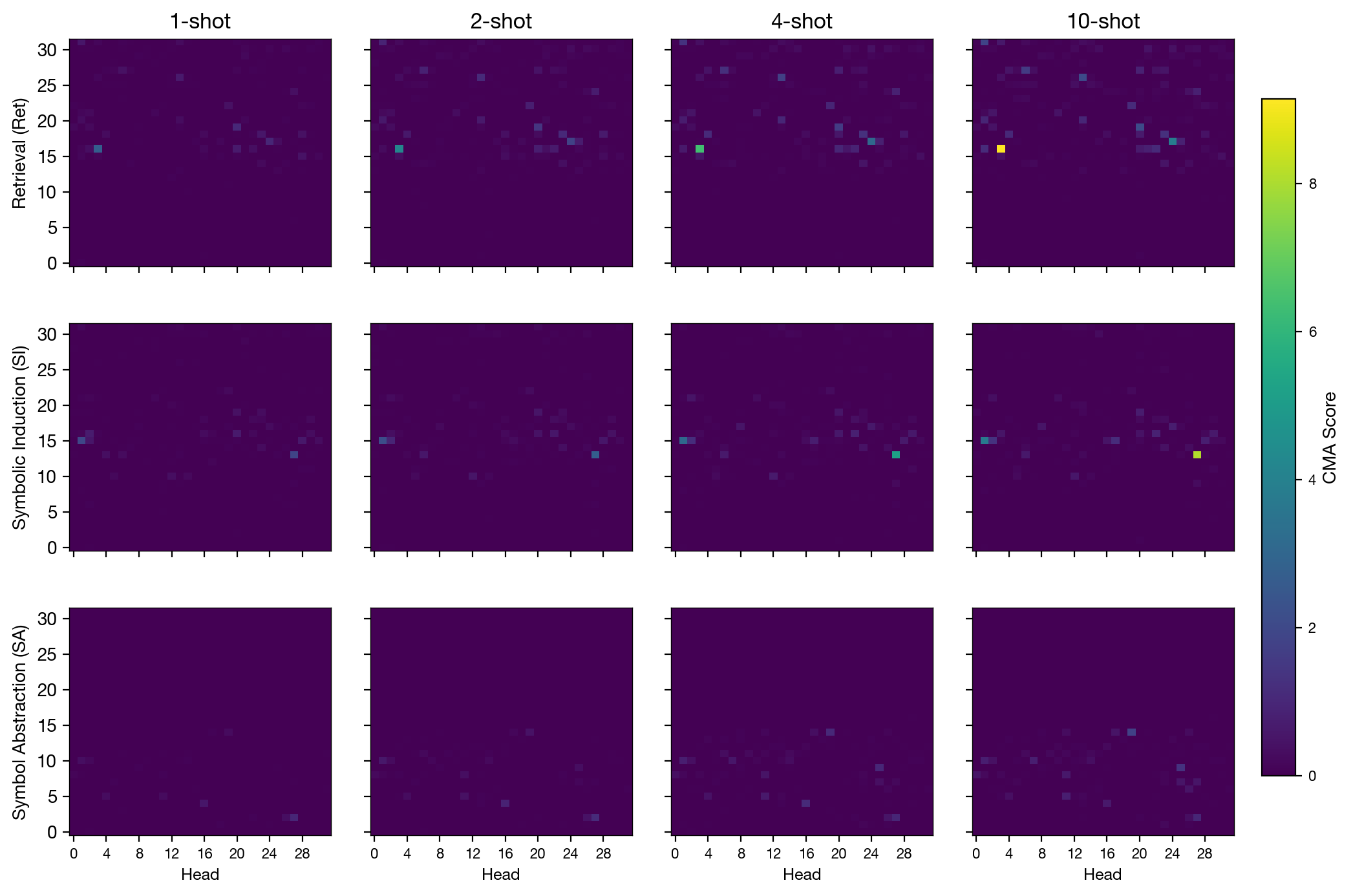}
    \caption{\textbf{The Symbolic Reasoning Circuit in Llama 3.1-8B at 1 through 10-Shot.} CMA Scores by layer by head, by circuit stage across shot counts.}
    \label{fig:circuit-by-shot-llama-3.1-8b}
\end{figure*}

The same three-stage topology recovered in Gemma is visible in Llama 3.1-8B (Figure~\ref{fig:circuit-by-shot-llama-3.1-8b}); persistent cores and RBO are reported in Section~\ref{sec:appendix-rbo}.

\subsubsection{Qwen 3-4B}
\begin{figure*}[t]
    \centering
    \includegraphics[width=\textwidth]{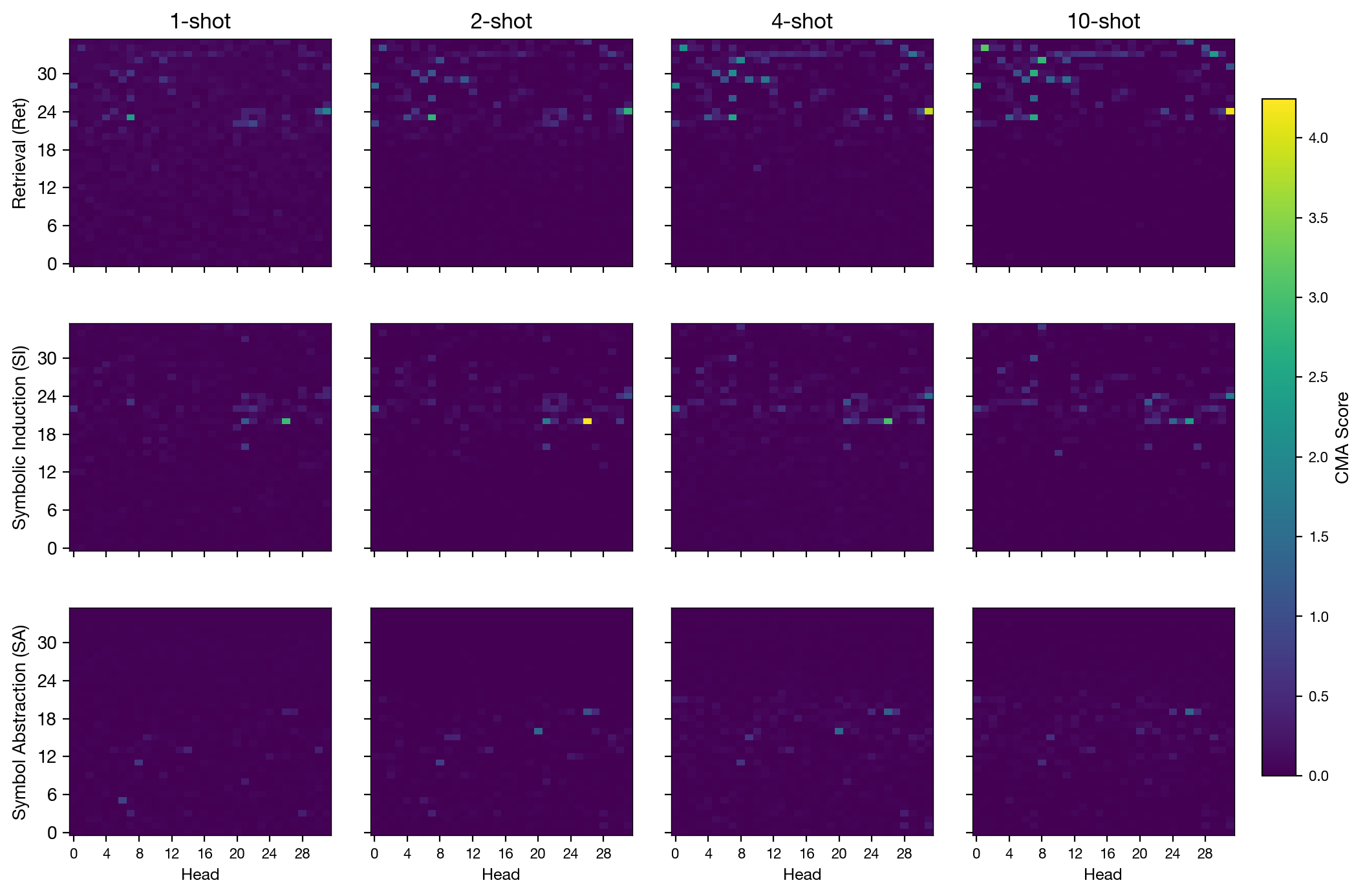}
    \caption{\textbf{The Symbolic Reasoning Circuit in Qwen 3-4B at 1 through 10-Shot.} CMA Scores by layer by head, by circuit stage across shot counts.}
    \label{fig:circuit-by-shot-qwen-3-4b}
\end{figure*}

Qwen 3-4B (Figure~\ref{fig:circuit-by-shot-qwen-3-4b}) shows the same topology with more reranking among the leading heads, especially at low shot counts and in SA, with some apparent diffusion of causal impact across heads at higher shot counts.

\subsubsection{Top-K heads, persistent cores, and rank-biased overlap}
\label{sec:appendix-rbo}

For each model and stage we average per-pair CMA scores over the 200 context pairs and over the two rule directions (ABA, ABB) to obtain a single mean CMA score per attention head, then rank all heads by descending score. Tables~\ref{tab:top5-heads}--\ref{tab:rbo} summarize the result.

\paragraph{Top-5 heads per stage and shot count.}
Table~\ref{tab:top5-heads} lists the top-5 heads at each shot count, with their mean CMA score in parentheses. The pattern is similar across models: the same heads dominate from 1-shot through 10-shot, and changes are dominated by score growth rather than identity. 

\begin{table}[!ht]
\centering
\scriptsize
\setlength{\tabcolsep}{4pt}
\begin{tabular}{ll ccccc}
\toprule
Stage & Shot & Top-5 (CMA score) \\
\midrule
\multicolumn{7}{l}{\textit{Gemma 2-2B}} \\
\midrule
\multirow{4}{*}{SA}
  & 1  & \textbf{L6H1} (0.50) & \textbf{L8H1} (0.45) & \textbf{L6H2} (0.38) & L3H1 (0.14)  & L3H0 (0.14) \\
  & 2  & \textbf{L6H1} (1.05) & \textbf{L8H1} (0.88) & \textbf{L6H2} (0.63) & L10H4 (0.47) & L3H1 (0.22) \\
  & 4  & \textbf{L6H1} (1.73) & \textbf{L8H1} (1.33) & L10H4 (1.01)         & \textbf{L6H2} (0.87) & L12H3 (0.58) \\
  & 10 & \textbf{L6H1} (2.07) & \textbf{L8H1} (2.03) & L10H4 (1.72)         & L12H3 (1.30) & \textbf{L6H2} (1.22) \\
\midrule
\multirow{4}{*}{SI}
  & 1  & \textbf{L14H0} (1.40) & \textbf{L12H1} (0.73) & \textbf{L15H0} (0.61) & \textbf{L14H1} (0.31) & L13H0 (0.25) \\
  & 2  & \textbf{L14H0} (2.73) & \textbf{L12H1} (1.58) & \textbf{L15H0} (0.90) & \textbf{L14H1} (0.76) & L17H3 (0.41) \\
  & 4  & \textbf{L14H0} (5.78) & \textbf{L12H1} (2.85) & \textbf{L14H1} (1.67) & \textbf{L15H0} (1.22) & L17H3 (0.99) \\
  & 10 & \textbf{L14H0} (9.02) & \textbf{L12H1} (3.66) & \textbf{L14H1} (2.71) & L17H3 (1.83)          & \textbf{L15H0} (1.50) \\
\midrule
\multirow{4}{*}{Ret}
  & 1  & \textbf{L18H6} (0.82) & \textbf{L22H4} (0.70) & L17H0 (0.66)         & L25H4 (0.41)          & \textbf{L20H7} (0.38) \\
  & 2  & \textbf{L18H6} (2.48) & \textbf{L22H4} (1.63) & L17H0 (1.04)         & \textbf{L20H7} (0.96) & L19H1 (0.70) \\
  & 4  & \textbf{L18H6} (3.94) & \textbf{L22H4} (2.55) & \textbf{L20H7} (1.49) & L17H0 (1.27)          & L19H1 (1.25) \\
  & 10 & \textbf{L18H6} (6.73) & \textbf{L22H4} (4.86) & L19H1 (2.38)         & \textbf{L20H7} (2.22) & L25H4 (1.69) \\
\midrule
\multicolumn{7}{l}{\textit{Llama 3.1-8B}} \\
\midrule
\multirow{4}{*}{SA}
  & 1  & \textbf{L2H27} (0.63)  & L4H16 (0.44)          & L5H11 (0.26)          & \textbf{L14H19} (0.25) & \textbf{L10H1} (0.24) \\
  & 2  & \textbf{L2H27} (0.76)  & L4H16 (0.58)          & \textbf{L10H1} (0.52) & \textbf{L14H19} (0.42) & L5H11 (0.35) \\
  & 4  & L9H25 (0.86)           & \textbf{L14H19} (0.86) & \textbf{L2H27} (0.85) & \textbf{L10H1} (0.75) & L4H16 (0.66) \\
  & 10 & \textbf{L14H19} (1.30) & L9H25 (1.13)          & \textbf{L2H27} (0.80) & \textbf{L10H1} (0.74) & L5H11 (0.55) \\
\midrule
\multirow{4}{*}{SI}
  & 1  & \textbf{L15H1} (1.81)  & \textbf{L13H27} (1.47) & \textbf{L15H2} (0.43) & \textbf{L15H28} (0.41) & L10H12 (0.34) \\
  & 2  & \textbf{L13H27} (2.55) & \textbf{L15H1} (2.04)  & \textbf{L15H2} (0.68) & \textbf{L15H28} (0.45) & L13H6 (0.43) \\
  & 4  & \textbf{L13H27} (4.60) & \textbf{L15H1} (2.71)  & \textbf{L15H2} (0.83) & \textbf{L15H28} (0.60) & L15H17 (0.54) \\
  & 10 & \textbf{L13H27} (6.39) & \textbf{L15H1} (3.03)  & \textbf{L15H2} (0.96) & L15H17 (0.85)          & \textbf{L15H28} (0.74) \\
\midrule
\multirow{4}{*}{Ret}
  & 1  & \textbf{L16H3} (2.58) & \textbf{L19H20} (0.84) & \textbf{L17H24} (0.55) & \textbf{L26H13} (0.47) & L31H1 (0.37) \\
  & 2  & \textbf{L16H3} (3.75) & \textbf{L17H24} (1.37) & \textbf{L19H20} (1.28) & \textbf{L26H13} (0.99) & L27H6 (0.79) \\
  & 4  & \textbf{L16H3} (5.42) & \textbf{L17H24} (2.52) & \textbf{L26H13} (1.61) & \textbf{L19H20} (1.42) & L27H6 (1.24) \\
  & 10 & \textbf{L16H3} (7.47) & \textbf{L17H24} (2.98) & \textbf{L26H13} (1.87) & L31H1 (1.84)          & \textbf{L19H20} (1.60) \\
\midrule
\multicolumn{7}{l}{\textit{Qwen 3-4B}} \\
\midrule
\multirow{4}{*}{SA}
  & 1  & L5H6 (0.50)           & L11H8 (0.39)          & L13H14 (0.27)         & L3H28 (0.26)          & \textbf{L19H26} (0.25) \\
  & 2  & L16H20 (1.05)         & \textbf{L19H26} (0.93) & L11H8 (0.52)         & L19H27 (0.37)         & L15H9 (0.34) \\
  & 4  & \textbf{L19H26} (1.23) & L16H20 (0.92)        & L15H9 (0.48)          & L19H27 (0.46)         & L19H24 (0.43) \\
  & 10 & \textbf{L19H26} (1.12) & L19H24 (0.51)        & L15H9 (0.41)          & L19H27 (0.38)         & L16H20 (0.31) \\
\midrule
\multirow{4}{*}{SI}
  & 1  & \textbf{L20H26} (2.20) & L20H21 (0.89)        & L16H21 (0.62)         & L22H0 (0.35)          & L20H22 (0.27) \\
  & 2  & \textbf{L20H26} (3.51) & L20H21 (1.36)        & L16H21 (0.65)         & L22H0 (0.57)          & L20H30 (0.54) \\
  & 4  & \textbf{L20H26} (2.59) & L20H21 (0.86)        & L22H0 (0.83)          & L24H31 (0.80)         & L23H21 (0.75) \\
  & 10 & \textbf{L20H26} (1.67) & L20H24 (1.09)        & L24H31 (0.93)         & L23H21 (0.82)         & L24H22 (0.58) \\
\midrule
\multirow{4}{*}{Ret}
  & 1  & \textbf{L23H7} (2.09) & \textbf{L24H31} (1.19) & L29H11 (0.64)        & \textbf{L28H0} (0.55) & L24H30 (0.52) \\
  & 2  & \textbf{L23H7} (2.71) & \textbf{L24H31} (2.18) & \textbf{L28H0} (1.35) & L29H11 (1.15)         & L34H1 (1.02) \\
  & 4  & \textbf{L24H31} (3.26) & \textbf{L23H7} (2.53) & L34H1 (2.12)         & \textbf{L28H0} (2.04) & L32H8 (1.74) \\
  & 10 & \textbf{L24H31} (3.69) & L34H1 (2.91)         & L32H8 (2.48)          & \textbf{L23H7} (2.44) & \textbf{L28H0} (2.23) \\
\bottomrule
\end{tabular}
\caption{Top-5 attention heads per stage and shot count, ranked left-to-right by mean CMA score (in parentheses; averaged across pairs and across ABA/ABB rule directions). \textbf{Bold} marks persistent-core heads (top-5 at every shot count). Reading down a column shows reranking among the leading heads.}
\label{tab:top5-heads}
\end{table}

\paragraph{Persistent cores.}
The \emph{persistent core} at top-$K$ is the set of heads that appear in the top-$K$ at every shot count $n \in \{1, 2, 4, 10\}$. Table~\ref{tab:persistent-core} reports the size of the persistent core (and the heads themselves at top-5) for each model and stage. Random-sample expected top-$K$ overlap is $K^2 / |H|$ where $|H|$ is the total head count: $0.12$ for Gemma top-5, $0.024$ for Llama, $0.022$ for Qwen.

\begin{table}[h]
\centering
\small
\begin{tabular}{lrrl}
\toprule
Stage & Top-5 core size & Top-10 core size & Top-5 core members \\
\midrule
\multicolumn{4}{l}{\textit{Gemma 2-2B}} \\
\midrule
SA  & 3 & 6 & L6H1, L6H2, L8H1 \\
SI  & 4 & 6 & L12H1, L14H0, L14H1, L15H0 \\
Ret & 3 & 7 & L18H6, L20H7, L22H4 \\
\midrule
\multicolumn{4}{l}{\textit{Llama 3.1-8B}} \\
\midrule
SA  & 3 & 5 & L2H27, L10H1, L14H19 \\
SI  & 4 & 6 & L13H27, L15H1, L15H2, L15H28 \\
Ret & 4 & 6 & L16H3, L17H24, L19H20, L26H13 \\
\midrule
\multicolumn{4}{l}{\textit{Qwen 3-4B}} \\
\midrule
SA  & 1 & 5 & L19H26 \\
SI  & 1 & 4 & L20H26 \\
Ret & 3 & 7 & L23H7, L24H31, L28H0 \\
\bottomrule
\end{tabular}
\caption{Persistent cores: heads in the top-$K$ at every shot count $n \in \{1, 2, 4, 10\}$. Qwen's top-5 is noisier at low shot counts but recovers at top-10.}
\label{tab:persistent-core}
\end{table}

\paragraph{Rank-Biased Overlap.}
RBO is a top-weighted similarity measure for ranked lists \citep{webber2010rbo} that downweights tail rankings, addressing the fact that $\sim$90\% of heads have near-zero CMA scores whose rank order is dominated by noise. We use the extrapolated form,
\begin{equation*}
\text{RBO}_{\text{ext}}(p) \;=\; (1 - p) \sum_{d=1}^{D} \frac{X_d}{d}\, p^{\,d-1} \;+\; p^{D}\, \frac{X_D}{D},
\end{equation*}
where $X_d$ is the size of the intersection of the two top-$d$ prefixes and $D = |H|$ is the full ranking depth, with persistence $p = 0.9$ (effective weight $\approx 86\%$ on the top 10 ranks). RBO ranges from 0 (disjoint) to 1 (identical). Table~\ref{tab:rbo} reports RBO across all pairs of shot counts.

\begin{table}[h]
\centering
\small
\begin{tabular}{lcccccc}
\toprule
& 1$\leftrightarrow$2 & 1$\leftrightarrow$4 & 1$\leftrightarrow$10 & 2$\leftrightarrow$4 & 2$\leftrightarrow$10 & 4$\leftrightarrow$10 \\
\midrule
\multicolumn{7}{l}{\textit{Gemma 2-2B}} \\
\midrule
SA  & 0.77 & 0.71 & 0.66 & 0.86 & 0.79 & 0.91 \\
SI  & 0.81 & 0.77 & 0.72 & 0.86 & 0.81 & 0.91 \\
Ret & 0.84 & 0.79 & 0.78 & 0.94 & 0.86 & 0.87 \\
\midrule
\multicolumn{7}{l}{\textit{Llama 3.1-8B}} \\
\midrule
SA  & 0.85 & 0.58 & 0.54 & 0.67 & 0.62 & 0.81 \\
SI  & 0.74 & 0.70 & 0.65 & 0.89 & 0.84 & 0.92 \\
Ret & 0.85 & 0.79 & 0.76 & 0.90 & 0.83 & 0.90 \\
\midrule
\multicolumn{7}{l}{\textit{Qwen 3-4B}} \\
\midrule
SA  & 0.41 & 0.28 & 0.29 & 0.67 & 0.55 & 0.80 \\
SI  & 0.84 & 0.68 & 0.45 & 0.80 & 0.55 & 0.68 \\
Ret & 0.85 & 0.69 & 0.57 & 0.77 & 0.64 & 0.83 \\
\bottomrule
\end{tabular}
\caption{Rank-Biased Overlap (RBO, $p=0.9$, full-depth extrapolated) between per-head CMA rankings at every pair of shot counts. Each ranking averages across the 200 CMA pairs and across the ABA and ABB rule directions. Higher is more similar; for full rankings of unrelated lists RBO would tend to $\approx 1 - p = 0.1$.}
\label{tab:rbo}
\end{table}

The pattern is consistent across the three models. RBO is high among shot counts where accuracy is non-trivial (e.g., $\geq 4$-shot for Gemma): adjacent-shot RBO sits at $0.81$--$0.94$ for SI/Ret in Gemma and Llama, and at $0.68$--$0.94$ for SI/Ret in Qwen. The lowest RBO values appear at 1-shot for the two models with the weakest 1-shot accuracy (Qwen at 27\% ABA, Gemma at 17\% ABA), and within those, primarily in the SA stage where the absolute CMA signal is also smallest, consistent with the SA stage being noisier when there are fewer in-context examples to abstract from. Even in those low-signal regimes the persistent-core counts and top-5 overlaps remain well above any random baseline.

Bootstrap resampling over the CMA pairs (1000 iterations, percentile 95\% CIs) confirms the RBO point estimates in Table~\ref{tab:rbo} are tight, e.g.\ Gemma SI $1\leftrightarrow 10$: $0.72$ $[0.65, 0.75]$; Qwen SI $1\leftrightarrow 10$: $0.45$ $[0.40, 0.51]$. A permutation test against independent head rankings (5000 iterations) places every observed RBO at $p<0.001$. 

\subsection{Extended Patching Results}

\begin{figure*}[t]
    \centering
    \includegraphics[width=\textwidth]{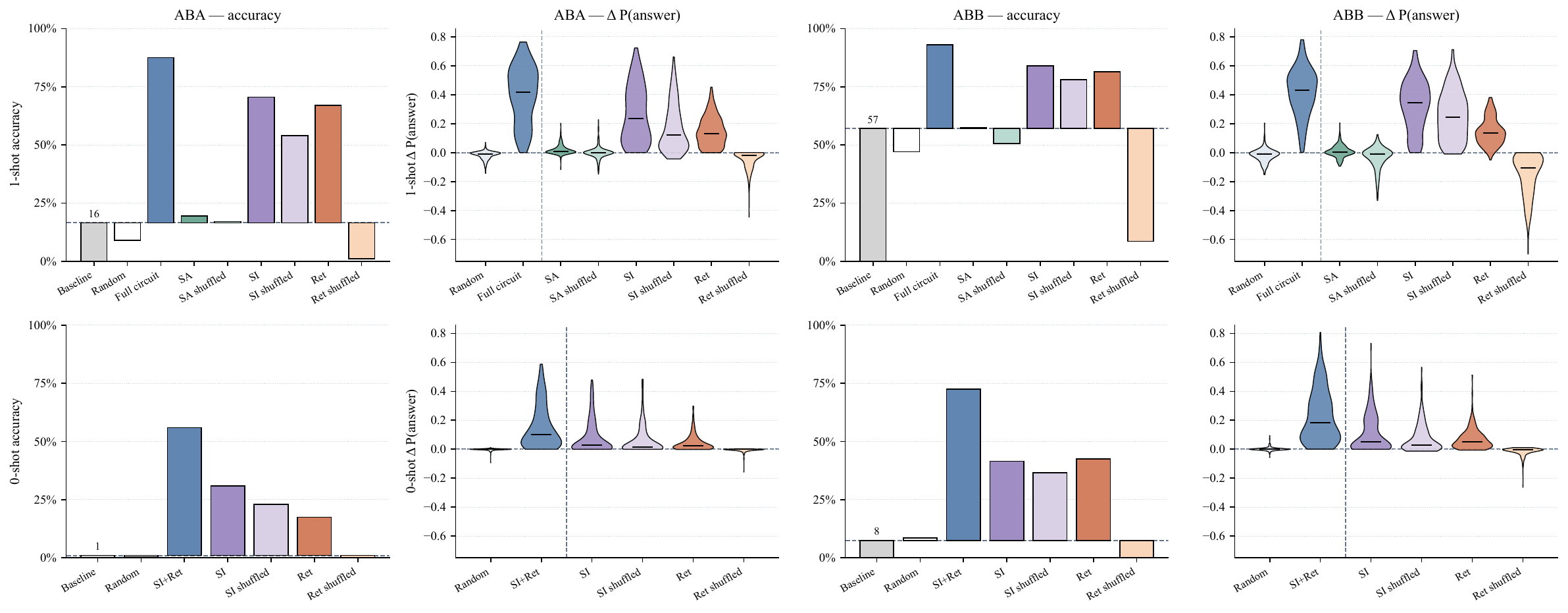}
    \caption{\textbf{Cross-shot patching, Gemma 2-2B.} Rows: 1-shot, 0-shot targets. Columns: ABA, ABB. Dashed line: baseline.}
    \label{fig:appendix_patching_gemma_2_2b}

    \vspace{1em}

    \includegraphics[width=\textwidth]{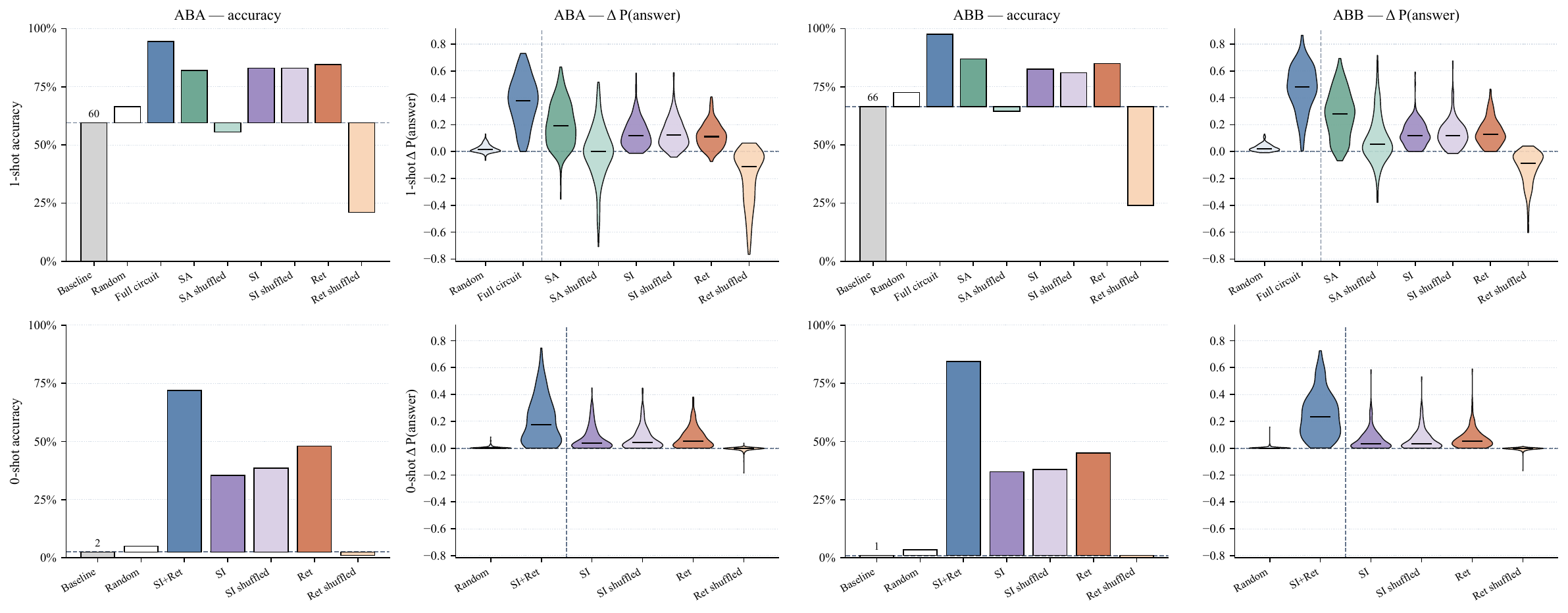}
    \caption{\textbf{Cross-shot patching, Llama 3.1-8B.} Rows: 1-shot, 0-shot targets. Columns: ABA, ABB. Dashed line: baseline.}
    \label{fig:appendix_patching_llama_31_8b}

    \vspace{1em}

    \includegraphics[width=\textwidth]{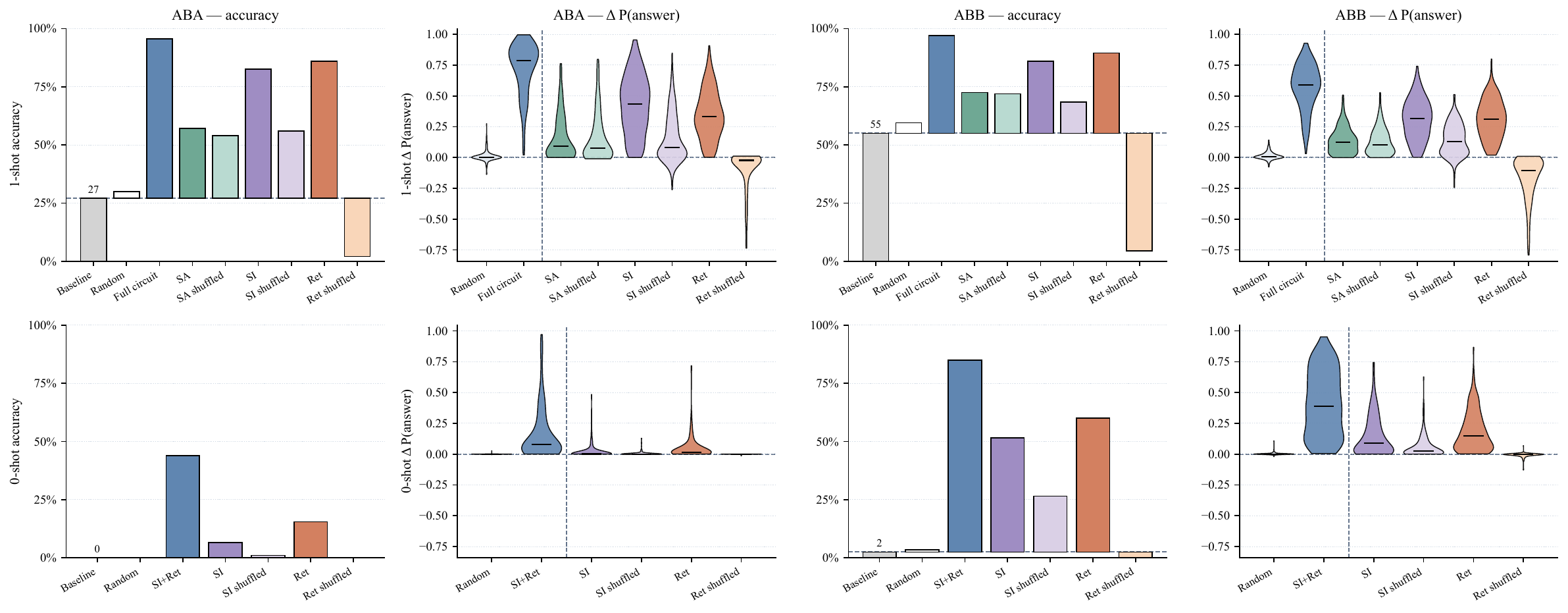}
    \caption{\textbf{Cross-shot patching, Qwen 3-4B.} Rows: 1-shot, 0-shot targets. Columns: ABA, ABB. Dashed line: baseline.}
    \label{fig:appendix_patching_qwen3_4b}
\end{figure*}

See Figure~\ref{fig:appendix_patching_gemma_2_2b}, \ref{fig:appendix_patching_llama_31_8b}, \ref{fig:appendix_patching_qwen3_4b}, and Table~\ref{tab:cross-shot-patching}.

\begin{table*}[h]
\centering
\small
\begin{tabular}{llccccccc}
\toprule
Shot & Rule & Baseline & SA-only & SI-only & SI-shuffled & Ret-only & All stages & Random \\
\midrule
\multicolumn{9}{l}{\textit{Gemma 2-2B}} \\
\midrule
0 & ABA & 1.0  & 1.0  & 31.0 & 23.0 & 17.5 & \textbf{56.0} & 0.5  \\
0 & ABB & 7.5  & 7.5  & 41.5 & 36.5 & 42.5 & \textbf{72.5} & 8.5  \\
1 & ABA & 16.5 & 19.5 & 70.5 & 54.0 & 67.0 & \textbf{87.5} & 9.0  \\
1 & ABB & 57.0 & 57.5 & 84.0 & 78.0 & 81.5 & \textbf{93.0} & 47.0 \\
\midrule
\multicolumn{9}{l}{\textit{Llama 3.1-8B}} \\
\midrule
0 & ABA & 2.5  & 2.5  & 35.5 & 38.5 & 48.0 & \textbf{72.0} & 5.0  \\
0 & ABB & 1.0  & 1.0  & 37.0 & 38.0 & 45.0 & \textbf{84.5} & 3.5  \\
1 & ABA & 59.5 & 82.0 & 83.0 & 83.0 & 84.5 & \textbf{94.5} & 66.5 \\
1 & ABB & 66.5 & 87.0 & 82.5 & 81.0 & 85.0 & \textbf{97.5} & 72.5 \\
\midrule
\multicolumn{9}{l}{\textit{Qwen 3-4B}} \\
\midrule
0 & ABA & 0.0  & 0.0  & 6.5  & 1.0  & 15.5 & \textbf{44.0} & 0.0  \\
0 & ABB & 2.5  & 2.5  & 51.5 & 26.5 & 60.0 & \textbf{85.0} & 3.5  \\
1 & ABA & 27.0 & 57.0 & 82.5 & 56.0 & 86.0 & \textbf{95.5} & 30.0 \\
1 & ABB & 55.0 & 72.5 & 86.0 & 68.5 & 89.5 & \textbf{97.0} & 59.5 \\
\bottomrule
\end{tabular}
\caption{\textbf{Cross-shot patching accuracy (\%).} Patching from a 10-shot donor into a 0/1-shot target, by stage. SA-only, SI-only, and Ret-only patch only the heads at that stage's positions; \emph{All stages} patches the union of all three. \emph{SI-shuffled} uses a donor with the same rule but entirely disjoint tokens, patched at SI heads only. \emph{Random} patches a matched-count random selection of non-circuit heads. $n{=}200$ prompts per cell. SA-only is identical to baseline at 0-shot because there are no in-context C positions to patch. Bold marks the All-stages column. Mean answer probability $p$ tracks accuracy and is omitted for compactness.}
\label{tab:cross-shot-patching}
\end{table*}

\subsection{Letter-String Analogies}
\label{appendix:lsa}

To check that the topology + amplification pattern is not specific to identity rules, we replicate the CMA analysis on the letter-string analogy task of \citet{webb2023emergent}, also used in \citet{yang2025emergentsymbolicmechanismssupport}. Two rules: \emph{successor} (\texttt{[a b c] [a b d]}, last letter $+1$) and \emph{predecessor} (\texttt{[b c d] [a c d]}, first letter $-1$). A prompt is scored correct only when all three answer letters are the argmax. We evaluate $500$ prompts per rule per shot count.

\paragraph{Accuracy.} Successor is largely solved by 2-shot in both models. Predecessor is harder and mirrors the ABA/ABB asymmetry. Per-letter inspection shows the transformed letter is the bottleneck (Gemma 2-shot predecessor letter 1: $23.6\%$; letter 3, unchanged: $100\%$).

\begin{table}[h]
\centering
\small
\begin{tabular}{llcc}
\toprule
Model & Shot & Succ. & Pred. \\
\midrule
Gemma 2-2B   & 2  & 89.0\% & 17.6\% \\
Gemma 2-2B   & 10 & 99.8\% & 76.4\% \\
Llama 3.1-8B & 2  & 98.4\% & 64.4\% \\
Llama 3.1-8B & 10 & 100\%  & 100\%  \\
\bottomrule
\end{tabular}
\caption{LSA accuracy (all three answer letters correct), 500 prompts per rule.}
\label{tab:lsa-acc}
\end{table}

\paragraph{CMA design.} SA and SI contrasts pair successor and predecessor prompts with identical starting letters; Ret contrasts pair same-rule prompts with different query letters. SA is patched at the output letters of in-context examples, SI and Ret at the positions preceding answer letters. Logit shifts are summed over rule-discriminating positions (letters 1 and 3 for SA/SI, all three for Ret). Filtered to correctly answered prompts, $200$ pairs per stage per condition.

\begin{figure}[h]
\centering 
\includegraphics[width=\textwidth]{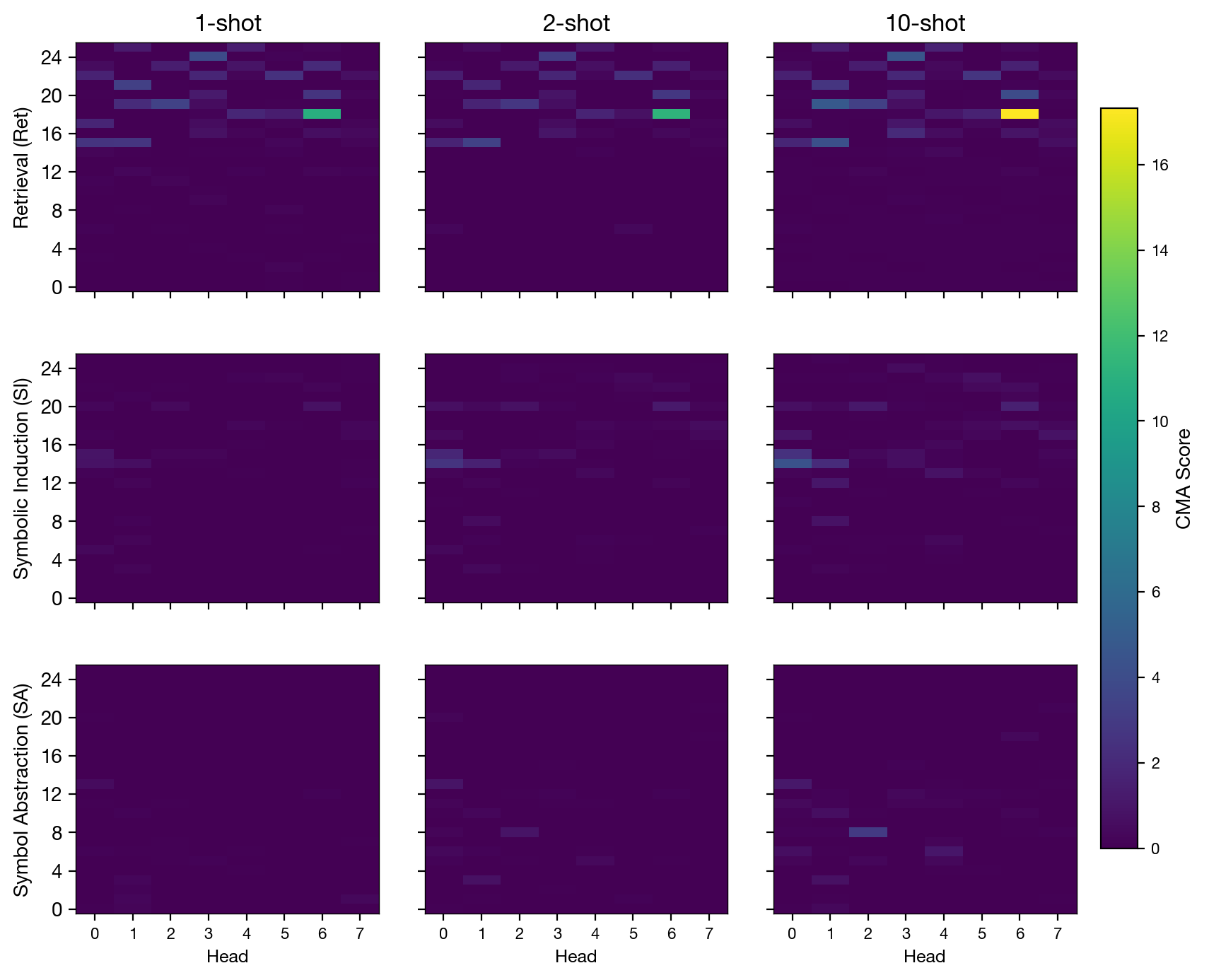}
\caption{\textbf{LSA CMA heatmaps, Gemma 2-2B} at 1, 2, and 10 shot, by stage. Same three-stage topology as ABA/ABB; L14H0 (SI) and L18H6 (Ret) dominate at all shot counts.}
\label{fig:lsa-cma}
\end{figure}

\paragraph{Results.} The same three-stage topology appears here (Figure~\ref{fig:lsa-cma}), but cross-task overlap of dominant heads is partial: L14H0 (SI) and L18H6 (Ret) lead in both task families and at every shot count, whereas the SA stage uses different heads across the two tasks. In LSA, the most persistent SA head is L13H0 (top-3 at 1, 2, and 10 shot), with L8H2 emerging as the SA leader by 10-shot ($0.05 \to 0.59 \to 2.39$).  Amplification replicates on all three stages of the cross-task heads: L14H0 (SI) grows $0.81 \to 2.20 \to 4.11$ (1/2/10-shot, $5.1\times$) and L18H6 (Ret) grows $10.6 \to 11.0 \to 17.3$ ($1.6\times$) and L13H0 (SA) $0.35 \to 0.62 \to 0.78$ ($2.2\times$).

\paragraph{Caveats.} Predecessor abstract pairs are sparse at low shot: at 1-shot the predecessor Ret condition has zero valid pairs and is omitted, and at 2-shot only $4$ pairs survive. Successor and 10-shot results are unaffected. We did not run LSA CMA on Llama or Qwen, nor extend cross-shot patching to LSA; this section is a generalization check on the CMA topology claim, not a full replication.

\end{document}